\documentclass[sigconf]{acmart}

\AtBeginDocument{%
  }

\usepackage{booktabs}
\usepackage{adjustbox}   
\usepackage{graphicx}
\usepackage{color}
\usepackage{subfig}
\usepackage[table]{xcolor}
\usepackage{booktabs}
\usepackage{multirow}
\usepackage{makecell}
\usepackage{xcolor}
\usepackage[dvipsnames]{xcolor}

\setcopyright{acmlicensed}
\copyrightyear{2026}
\acmYear{2026}
\setcopyright{cc}
\setcctype{by-nc-nd}
\setcctype{by}

\acmDOI{10.1145/3767308.3836233}
\acmConference[MM '26] {Proceedings of the 34th ACM International Conference on Multimedia}{November 10--14, 2026}{Rio de Janeiro, Brazil.}
\acmBooktitle{Proceedings of the 34th ACM International Conference on Multimedia (MM '26), November 10--14, 2026, Rio de Janeiro, Brazil}
\acmISBN{979-8-4007-2213-4/2026/11}

\begin{document}

\title[Leveraging 2D Foundation Priors for GCD-TTA in 3D LiDAR Segmentation]{Leveraging 2D Foundation Priors for Generalized Category Discovery under Test-Time Adaptation in 3D LiDAR Segmentation}

\author{Hanqi Chen}
\authornote{These authors are with the Guangdong Provincial Key Laboratory of Visual Media and Multidimensional Intelligence.}
\affiliation{%
  \institution{CSSE, Shenzhen University}
  \city{Shenzhen}
  \state{Guangdong}
  \country{China}}
\email{chenhanqi0613@gmail.com}

\author{Chaoqi Chen}
\authornotemark[1]
\authornote{Corresponding author.}
\affiliation{%
  \institution{CSSE, Shenzhen University}
  \city{Shenzhen}
  \state{Guangdong}
  \country{China}}
\email{cqchen1994@gmail.com}

\author{Tian Ji}
\affiliation{%
 \institution{The Hong Kong University of Science and Technology (Guangzhou)}
 \city{Guangzhou}
 \state{Guangdong}
 \country{China}}
\email{tji549@connect.hkust-gz.edu.cn}

\author{Hui Huang}
\authornotemark[1]
\affiliation{%
  \institution{CSSE, Shenzhen University}
  \city{Shenzhen}
  \state{Guangdong}
  \country{China}}
\email{hhzhiyan@gmail.com}

\begin{abstract}

In autonomous driving, 3D LiDAR semantic segmentation is expected to adapt to evolving environments and recognize previously unseen categories, making generalized category discovery under test-time adaptation (GCD-TTA) an important problem. However, existing works treat GCD and TTA separately, while the limited semantic information in 3D point clouds makes GCD-TTA under domain shift more prone to feature collapse when combining GCD with TTA. To address this issue, we propose a two-stage framework for GCD-TTA in 3D LiDAR semantic segmentation. Specifically, in the first stage, we adaptively aggregate local 3D features according to geometric density variations and obtain reliable pseudo labels under a global class-wise prototype consistency constraint. In the second stage, we cast category discovery as an optimal transport assignment process for globally consistent grouping of novel categories. Meanwhile, semantic features from 2D vision foundation models are fused with 3D point cloud features to enrich 3D semantic information and promote a more separable feature space. Extensive experiments on Synth4D to SemanticKITTI, SynLiDAR to SemanticKITTI, and Synth4D to nuScenes demonstrate the effectiveness of our framework. Under a unified evaluation protocol, our method achieves the best overall mIoU on 9 out of 10 benchmark splits and improves the overall mIoU by up to 12.21 points over the strongest re-implemented baseline. %

\end{abstract}




\maketitle

\section{Introduction}

\begin{figure}[!t]
    \centering
    \includegraphics[width=\linewidth]{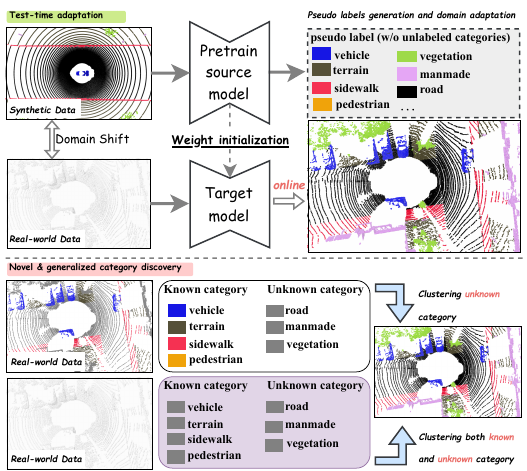}
    \caption{Illustration of the TTA and GCD pipelines.}
    \label{fig:gcd-tta_comp}
\end{figure}

The semantic segmentation of LiDAR point clouds is fundamental to autonomous driving, providing fine-grained scene understanding for downstream tasks such as planning and safe navigation~\cite{Automotive_LiDAR_seg}. 
Beyond safety requirements~\cite{xi2018detection}, real-world deployment faces two coupled challenges: domain shifts caused by changing data distributions~\cite{unimix} and the emergence of unseen semantic categories in open-world scenarios~\cite{On-the-Fly}. 
Therefore, practical 3D LiDAR segmentation models should generalize across environments and discover categories absent from the source domain, corresponding to test-time adaptation (TTA)~\cite{tta_survey} and generalized category discovery (GCD)~\cite{gcd}, respectively.

Existing 3D LiDAR TTA methods~\cite{GIPSO,HGL,good,D3CTTADD,ATTA} mainly follow a synthetic-to-real paradigm, adapting models pretrained on labeled synthetic domains (e.g., SynLiDAR~\cite{SynLiDAR}, Synth4D~\cite{GIPSO}) to unlabeled real-world data (e.g., SemanticKITTI~\cite{semantickitti}, nuScenes~\cite{nuscenes}). 
Meanwhile, although GCD has been extensively studied in 2D vision~\cite{tang2026bures,tang2025gcd,SimGCD,active_gcd,gcd,Semantic_Segmentation_gcd,Rathore2025DGCD}, most 3D approaches~\cite{NCD,SNOPS,DASL,CaRR} focus on novel class discovery (NCD), assuming that only unseen categories require grouping. 
Recent work~\cite{Kim_2026_WACV} extends 3D LiDAR segmentation toward GCD by considering disjoint labeled and unlabeled sets containing both known and novel categories. 
However, existing studies still treat TTA and GCD as separate problems, leaving their joint challenge unexplored. Fig.~\ref{fig:gcd-tta_comp} illustrates their distinct pipelines.

Jointly addressing TTA and GCD is particularly challenging due to domain shift, online unsupervised adaptation, and the sparse, ambiguous nature of 3D point clouds. These factors can destabilize feature representations and cause feature collapse during adaptation. Consequently, noisy pseudo labels induce confirmation bias, where repeated self-supervision reinforces incorrect alignment and category assignments, ultimately degrading class separability and representation capacity.

\begin{figure*}
    \centering
    \includegraphics[width=\linewidth]{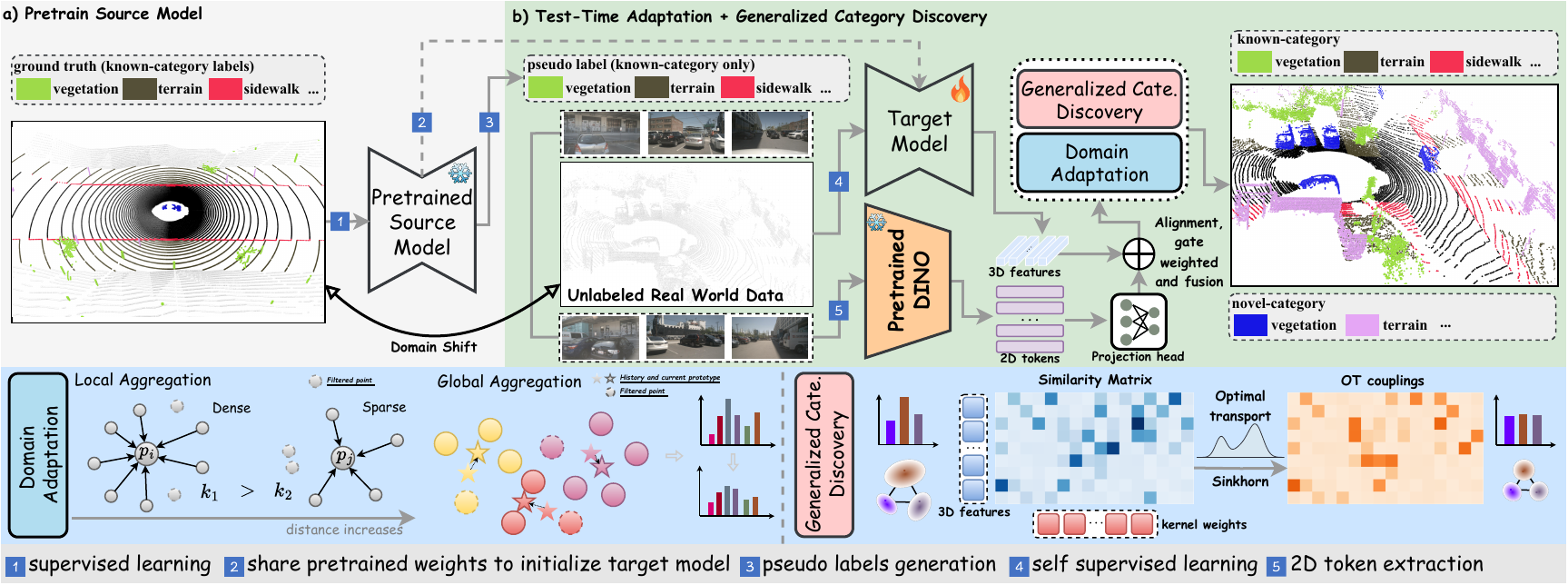}
    \caption{Overview of the source pretraining, test-time adaptation, and generalized category discovery pipeline.}
    \label{fig:pipeline}
\end{figure*}

To address these challenges, we introduce generalized category discovery under test-time adaptation (GCD-TTA) for 3D LiDAR semantic segmentation, aiming at unified adaptation and discovery. 
We adopt a teacher-student framework for online adaptation, as illustrated in Fig.~\ref{fig:pipeline}. 
For source-known categories, we refine teacher-generated pseudo labels using geometry priors to reduce domain-induced errors. 
For source-unknown categories, we employ Sinkhorn-based balanced assignments~\cite{cuturi2013sinkhorn} to discover emerging categories from unlabeled target points. 
Furthermore, we leverage transferable semantic priors from 2D foundation models (e.g., DINO~\cite{caron2021emerging,dinov2,simeoni2025dinov3}) to enhance feature discrimination and alleviate confirmation bias caused by noisy supervision. 
The proposed framework enables simultaneous target-domain adaptation and novel-category discovery.

Our contributions are summarized as follows:
\begin{itemize}
    \item We introduce generalized category discovery under test-time adaptation (GCD-TTA) for 3D LiDAR semantic segmentation and present the first unified framework for this setting.
    \item We propose a geometry-aware pseudo-label refinement strategy that integrates distance-aware filtering and global prototype consistency to suppress noisy supervision.
    \item We incorporate 2D foundation model features as transferable semantic priors, improving ambiguous-point discrimination and novel-category discovery while preserving shared-category performance.
\end{itemize}

\section{Related Work}

\subsection{Test-Time Adaptation}
Test-time adaptation (TTA) adapts a source-pretrained model to unlabeled target data at inference time, without accessing source-domain samples during deployment~\cite{tta_survey}. In existing 3D LiDAR semantic segmentation studies, GIPSO~\cite{GIPSO} formulates source-free online adaptation through adaptive self-training and geometric-feature propagation. MM-TTA~\cite{Shin_2022_CVPR} extends this setting to multimodal segmentation with intra-modal pseudo-label generation and inter-modal refinement. HGL~\cite{HGL} exploits hierarchical geometric structure through local pseudo-label generation, global prototype refinement, and temporal consistency. More recent methods move toward continual and deployment-oriented adaptation, such as D$^3$CTTA~\cite{D3CTTADD} and APCoTTA~\cite{APCoTTA}. GOOD~\cite{good} further highlights the importance of handling semantic shifts beyond the standard closed-set setting. Annotator~\cite{xie2023annotator} and ATTA~\cite{ATTA} incorporate active learning into TTA, leading to markedly improved adaptation efficiency. 

\subsection{Generalized Category Discovery}
Generalized category discovery (GCD) considers a realistic mixed setting with both known and novel categories~\cite{gcd}. In 2D vision, SimGCD~\cite{SimGCD} and PIM~\cite{PIM} show that parametric classification can be effective for this task, while CMS~\cite{CMS} improves clustering quality with mean-shift updates and contrastive learning. More practical variants have also been explored, including active GCD~\cite{active_gcd}, GCD for semantic segmentation~\cite{Semantic_Segmentation_gcd}, and related settings under distribution shifts~\cite{yi2021completelabeldomain}. 
In 3D LiDAR semantic segmentation, recent works~\cite{NOPS,SNOPS,DASL,CaRR} focus on novel category discovery, assuming coexisting labeled and unlabeled points in each scan, with unlabeled points belonging only to novel categories. Study~\cite{Kim_2026_WACV} further considers a generalized setting, where unlabeled data contain both known and novel categories and are disjoint from labeled data. This setting is more general and better reflects realistic open-world scenarios.

\section{Method}

\subsection{Problem Formulation}
Let the labeled source domain be denoted by $\mathcal{D}_{\mathcal S}=\{(P_i^{\mathcal S}, Y_i^{\mathcal S})\}_{i=1}^{N_{\mathcal S}}$, where $P_i^{\mathcal S}=\{p_n \mid p_n=(x_n,y_n,z_n)\}_{n=1}^{M_i}$ is a source-domain point cloud and $Y_i^{\mathcal S} = \{c_n \mid c_n \in \mathbb{N}^+\}_{n=1}^{M_i}$ is its class labels.
Following the generalized category discovery setting~\cite{NOPS}, the whole category space is divided into two disjoint subsets, namely the known category set $\mathcal{C}_{\mathrm{kn}} = \{1, 2, \dots, K\}$ and the novel category set $\mathcal{C}_{\mathrm{nov}} = \{K+1, K+2, \dots, C\}$, where $\mathcal{C}_{\mathrm{kn}} \cap \mathcal{C}_{\mathrm{nov}} = \varnothing$. During source pre-training, only labeled samples from known classes are used, i.e., $\mathcal{D}_\mathcal{S}^{\mathrm{kn}} = \{(P_i^\mathcal{S}, Y_i^\mathcal{S}) \mid \forall c_n \in Y_i^\mathcal{S}, \; c_n \in \mathcal{C}_{kn} \}$, based on which a source model $F_{\mathcal{S}}$ is trained. 

The target domain is an unlabeled LiDAR point cloud stream denoted by $\mathcal{D}_\mathcal{T} = \{P_t^\mathcal{T}\}_{t=1}^{T}$, where samples arrive sequentially in an online manner. Each target sample has an underlying but unavailable label belonging to the target category space $\mathcal{C}_\mathcal{T} = \mathcal{C}_{\mathrm{kn}} \cup \mathcal{C}_{\mathrm{nov}}$. In other words, the target stream contains both known and novel categories, while no target annotations are accessible during adaptation training. 

\subsection{Domain Adaptation}
Domain adaptation fundamentally relies on exploiting transferable priors that remain stable across domains, rather than relying on target-domain annotations \cite{phung2021learning}. This is especially important in GCD-TTA, where the key challenge is not simply to adapt to the target distribution, but to uncover domain-invariant cues that remain reliable under domain shift.

In LiDAR-based perception, such cues arise from two naturally complementary sources: 3D geometry and 2D semantics. Geometric priors originate from the inherent characteristics of LiDAR sensing, including distance-dependent density variation and spatial structure~\cite{hu2022point}, both of which provide robust discrimination cues. Semantic priors, by contrast, are provided by large-scale pre-trained visual models, which offer strong generalization and high-level semantic guidance~\cite{zhang2023lidar}. Together, these priors combine stable geometric regularity with rich semantic knowledge. Our method is motivated by this complementarity and exploits both to tackle the GCD-TTA problem.

\noindent \textbf{Geometry-aware local aggregation.}
For each target point $p_n^t$, the source model produces logits $\mathbf z_n^t = F_{\mathcal S}(p_n^t)$, from which we obtain the class distribution $\mathbf s_n^t=\operatorname{softmax}(\mathbf z_n^t/\tau)$ and the point-wise prediction $\hat y_n^t=\arg\max_{c\in\mathcal C_{\mathrm{kn}}}\mathbf s_n^t(c)$. Rather than directly accepting $\hat y_n^t$ as a pseudo label, we then estimate reliability of $\hat y_n^t$ using local geometric context.

As LiDAR point density decreases with distance, we adapt the neighborhood size according to the radial distance $r_n^t=\|p_n^t\|_2$:
\begin{equation}
k_n^t=\operatorname{clip}\!\left(
\left\lfloor
k_{\mathrm{ref}}\left(\frac{r_{\mathrm{ref}}}{r_n^t+\epsilon}\right)^\alpha
\right\rfloor,
k_{\min},k_{\max}
\right),
\end{equation}
where $k_{\mathrm{ref}}$ denotes the reference neighborhood size at distance $r_{\mathrm{ref}}$, $\alpha$ controls the distance sensitivity, and $k_{\min}$ and $k_{\max}$ prevent overly small or overly large neighborhoods. Unless otherwise specified, we set $\alpha = 1.0$, $k_{\mathrm{ref}} = 16$, $r_{\mathrm{ref}} = 20$, $k_{\min} = 8$, $k_{\max} = 24$, and $\epsilon$ = $1 \times 10^{-6}$ by default. We then construct the corresponding $k_n^t$-NN set $\mathcal N_n^t$ and aggregate neighboring predictions with distance-aware weights $w_{nm}^t \propto \exp(-\|p_n^t-p_m^t\|_2)$, normalized within $\mathcal N_n^t$:
\begin{equation}
\bar{\mathbf s}_n^t=\sum_{p_m^t\in\mathcal N_n^t} w_{nm}^t\,\mathbf s_m^t.
\end{equation}
The aggregated hard label is defined as $\bar y_n^t=\arg\max_c \bar{\mathbf s}_n^t(c)$.

We evaluate local reliability from two complementary cues. The first is measured by the normalized entropy of the aggregated distribution $\bar{\mathbf s}_n^t$. The second is measured by the normalized entropy of the neighborhood label histogram $\mathbf q_n^t$, where $q_n^t(c)=\frac{1}{|\mathcal N_n^t|}\sum_{p_m^t\in\mathcal N_n^t}\mathbf{1}(\bar y_m^t=c)$, and $\mathbf{1}(\cdot)$ denotes the indicator function. The final reliability score is
\begin{equation}
\psi_n^t=
\left(1-\frac{\mathcal H(\bar{\mathbf s}_n^t)}{\log K}\right)
\left(1-\frac{\mathcal H(\mathbf q_n^t)}{\log K}\right),
\end{equation}
where $\mathcal H(\mathbf p)=-\sum_{c=1}^{K} p(c)\log p(c)$ denotes the Shannon entropy. 

To avoid pseudo-label selection being dominated by frequent classes, we use class-wise reliability thresholds. For each predicted class $c\in\mathcal C_{\mathrm{kn}}$, we compute $\delta_c$ as the $\rho$-quantile of $\{\psi_n^t\mid \hat y_n^t=c\}$ and retain only points whose reliability exceeds the threshold of their predicted class:
\begin{equation}
\tilde y_n^{t,\mathrm{loc}}=
\begin{cases}
\hat y_n^t, & \psi_n^t>\delta_{\hat y_n^t},\\
\varnothing, & \text{otherwise}.
\end{cases}
\end{equation}
Where $\delta_c$ is set to 0.6 by default.

Unlike conventional pseudo-label filtering based only on point-wise confidence or fixed-size smoothing, our method adapts the aggregation range to the distance-dependent sampling pattern of LiDAR and combines prediction certainty with neighborhood semantic consistency. As a result, it suppresses isolated overconfident errors and yields cleaner known-class pseudo labels under domain shift.

\noindent\textbf{Prototype-based global aggregation.}
Although geometry-aware local aggregation suppresses isolated noisy predictions, it remains a local mechanism and may fail when an entire neighborhood is jointly corrupted by domain shift. To provide a more stable class-level prior, we further introduce a prototype-based global aggregation module over the target stream, as in~\cite{HGL}.

We decompose the source model as $F_{\mathcal S}=H_{\mathcal S}\circ E_{\mathcal S}$, where $E_{\mathcal S}$ is the feature extractor and $H_{\mathcal S}$ is the classifier over the known label space $\mathcal C_{\mathrm{kn}}$. For each target point $p_n^t$, we extract its source feature $\mathbf f_n^t = E_{\mathcal S}(p_n^t)\in\mathbb R^d$. We then maintain an online prototype bank for known classes. Optionally, prototypes can be partitioned by range bins, following the strategy in~\cite{D3CTTADD}. Specifically, each point is assigned a bin index $b(n)\in\{1,\dots,B\}$, where $B=1$ in the default setting without range partition.

Using the locally reliable pseudo labels $\tilde y_n^{t,\mathrm{loc}}$, we collect the target features assigned to class $c$ in bin $b$, denoted by $\mathcal S_{c,b}^t=\{n\mid \tilde y_n^{t,\mathrm{loc}}=c,\ b(n)=b\}$, and update the corresponding prototype by momentum averaging:
\begin{equation}
\mathbf{m}_{c,b}
\leftarrow
\mu\,\mathbf{m}_{c,b}
+
(1-\mu)\frac{1}{|\mathcal S_{c,b}^t|}
\sum_{n\in \mathcal S_{c,b}^t}\mathbf f_n^t,
\quad \text{if} \ \
|\mathcal S_{c,b}^t|>0.
\end{equation}
Here, $\mu$ is the momentum coefficient and we set it to 0.99. When some bins contain too few observations, we use the pooled prototype of class $c$ over available bins as a fallback to avoid unstable matching.

Given the prototype bank, we compute a global class score for each target point by cosine similarity:
\begin{equation}
a_{n,c}^t=
\left\langle
\frac{\mathbf f_n^t}{\|\mathbf f_n^t\|_2},
\frac{\mathbf m_{c,b(n)}}{\|\mathbf m_{c,b(n)}\|_2}
\right\rangle,
\quad c\in\mathcal C_{\mathrm{kn}}.
\end{equation}
The prototype-based global prediction is then obtained by selecting the class with the maximum similarity, i.e., $\tilde y_n^{t,\mathrm{glob}}=\arg\max_{c\in\mathcal C_{\mathrm{kn}}} a_{n,c}^t$.

Finally, we accept a known-class pseudo label only when the local and global predictions agree:
\begin{equation}
\tilde y_n^t=
\begin{cases}
\tilde y_n^{t,\mathrm{loc}}, & \tilde y_n^{t,\mathrm{loc}}=\tilde y_n^{t,\mathrm{glob}},\\
\varnothing, & \text{otherwise}.
\end{cases}
\end{equation}

Compared with purely local filtering, prototype-based global aggregation provides an explicit class-level memory over the target stream. Since the prototypes are updated online from reliable target points rather than fixed source statistics, they gradually adapt to the target distribution while still remaining anchored to the source semantic space. The agreement between local geometric consistency and global prototype matching therefore acts as an effective dual-criterion filter, yielding cleaner known-class pseudo labels under domain shift.

\noindent \textbf{Confidence-aware 2D--3D feature fusion.}
For each target point cloud $P_t^\mathcal{T}$, we further exploit its synchronized image observations $\mathcal I_t=\{I_{t,v}\}_{v=1}^{V_t}$ to provide semantic cues complementary to sparse LiDAR geometry. A frozen pretrained DINO encoder $\Phi_{\mathrm{2D}}$ extracts dense image descriptors. For each target point $p_n^t$, we project it onto the image plane and sample the corresponding descriptor from a valid visible view, which is then mapped into the 3D embedding space to obtain $\hat{\mathbf u}_n^t$. If no valid image observation is available, we set the visibility indicator $m_n^t=0$ and $\hat{\mathbf u}_n^t=\mathbf 0$. Meanwhile, the 3D branch produces a LiDAR feature $\mathbf f_n^{t,\mathrm{3D}}\in\mathbb R^d$.

To avoid indiscriminate cross-modal fusion, we estimate a point-wise confidence-aware gate that modulates the contribution of the 2D semantic cue:
\begin{equation}
g_n^t
=
m_n^t\,
\sigma\!\left(
G_{\mathrm{gate}}\!\left(
[\mathbf f_n^{t,\mathrm{3D}};\hat{\mathbf u}_n^t;\max_{c\in\mathcal C_{\mathrm{kn}}}\mathbf s_n^t(c)]
\right)
\right),
\end{equation}
where $G_{\mathrm{gate}}$ is a lightweight gating network and $\mathbf s_n^t$ denotes the source-model prediction distribution. The final fused representation is computed as
\begin{equation}
\tilde{\mathbf f}_n^t
=
\mathbf f_n^{t,\mathrm{3D}}
+
\lambda\, g_n^t\, \hat{\mathbf u}_n^t,
\end{equation}
where $\lambda$ controls the overall strength of the fusion and we set 0.6 by default.

This formulation treats the image branch as a semantic correction signal rather than a direct replacement for LiDAR features. The visibility term suppresses invalid cross-modal transfer, while the confidence-aware gate adaptively injects 2D semantics according to the reliability of the 3D prediction. In this way, the fused representation particularly benefits sparse, distant, and semantically ambiguous target points while reducing negative transfer from unreliable image regions.

\subsection{OT-based Category Discovery}

Given the target novel-point set $\mathcal U_t=\{p_n^t\}_{n=1}^{N_t}$, we adapt a target model $F_{\mathcal T}$, initialized from the source model $F_{\mathcal S}$, to partition these points into $C_{\mathrm{nov}}=|\mathcal C_{\mathrm{nov}}|=C-K$ novel categories. Let $E_{\mathcal T}$ denote the target encoder and $\mathbf g_n^t=E_{\mathcal T}(p_n^t)\in\mathbb R^d$ the feature of point $p_n^t$. To improve robustness, $F_{\mathcal T}$ is equipped with $H$ parallel discovery heads. The $h$-th head is parameterized by a prototype matrix $W_h=[\mathbf w_{h,1},\dots,\mathbf w_{h,C_{\mathrm{nov}}}] \in \mathbb R^{d\times C_{\mathrm{nov}}}$ and produces cosine-similarity logits
\begin{equation}
z_{n,c}^{t,h}
=
\left\langle
\frac{\mathbf g_n^t}{\|\mathbf g_n^t\|_2},
\frac{\mathbf w_{h,c}}{\|\mathbf w_{h,c}\|_2}
\right\rangle,
\quad c=1,\dots,C_{\mathrm{nov}}.
\end{equation}

Rather than assigning each point independently by a local $\arg\max$, we formulate novel-category discovery as a balanced entropy-regularized optimal transport problem. Let $\mathbf Z_h^t\in\mathbb R^{N_t\times C_{\mathrm{nov}}}$ collect the logits of the $h$-th head. We seek an assignment matrix $\mathbf Q_h^t\in\mathbb R_{+}^{N_t\times C_{\mathrm{nov}}}$ by
\begin{equation}
\begin{aligned}
\mathbf Q_h^{t\star}
&=
\arg\max_{\mathbf Q\in\mathcal Q}
\ \langle \mathbf Q,\mathbf Z_h^t \rangle
+
\varepsilon \mathcal H(\mathbf Q),
\\
\mathcal Q
&=
\left\{
\mathbf Q\ge 0\ \middle|\ 
\mathbf Q\mathbf 1_{C_{\mathrm{nov}}}=\mathbf 1_{N_t},\ 
\mathbf Q^\top\mathbf 1_{N_t}
=
\frac{N_t}{C_{\mathrm{nov}}}\mathbf 1_{C_{\mathrm{nov}}}
\right\},
\end{aligned}
\end{equation}
where $\mathcal H(\mathbf Q)=-\sum_{n,c}Q_{n,c}\log Q_{n,c}$ is the entropy regularizer. The row constraint yields a valid point-wise assignment, while the column constraint discourages collapse by encouraging balanced mass across categories. The resulting problem is solved efficiently by Sinkhorn--Knopp~\cite{cuturi2013sinkhorn,altschuler2017near} iterations. Compared with per-point hard assignment or unconstrained clustering, OT explicitly suppresses degenerate solutions in which most points collapse into a few dominant categories.

To further stabilize online discovery over the target stream, following~\cite{NOPS}, we maintain a memory queue of novel-point features from recent target samples and append them to the current batch during OT. This enlarges the transport set and makes the estimated marginals less sensitive to the statistics of a single frame. We further employ an overclustering branch with $\gamma C_{\mathrm{nov}}$ prototypes, where $\gamma>1$, so that fine-grained target variations can be preserved before being consolidated into semantic novel categories.

\subsection{Optimization}

Following the pseudo-label refinement and OT-based category discovery described above, we optimize the target model by bidirectional cross-view self-training. For each target point cloud $P_t^\mathcal{T}$, we generate two augmented views, denoted by $P_t^{\mathcal{T},1}$ and $P_t^{\mathcal{T},2}$. After voxelization and point sampling, only the points preserved in both views are used for supervision, and we denote this shared set by $\Omega_t$.

For notational simplicity, let $\mathbf z_{v,n}^{t,h}\in\mathbb R^{C}$ denote the full prediction logits of point $p_n^t$ in view $v\in\{1,2\}$ under the $h$-th discovery head, obtained by combining the source-initialized known-class classifier with the corresponding novel-category discovery head. The pseudo target $\mathbf q_{v,n}^{t,h}$ is defined on the same label space. For reliable known points, it is the one-hot encoding of $\tilde y_n^t$. For novel points, its novel-category part is given by the OT assignment and its known-category part is set to zero. We then optimize the model with standard soft-target cross-entropy in a bidirectional cross-view manner:
\begin{equation}
\mathcal{L}_{\mathrm{ce}}
=
\frac{1}{H}\sum_{h=1}^{H}
\frac{1}{|\Omega_t|}
\sum_{n\in\Omega_t}
\Big[
\ell_{\mathrm{ce}}(\mathbf{z}_{1,n}^{t,h}, \mathbf{q}_{2,n}^{t,h})
+
\ell_{\mathrm{ce}}(\mathbf{z}_{2,n}^{t,h}, \mathbf{q}_{1,n}^{t,h})
\Big],
\end{equation}
where $\ell_{\mathrm{ce}}(\mathbf z,\mathbf q)=-\sum_c q(c)\log(\operatorname{softmax}(\mathbf z)(c))$.

To capture finer target structure, we additionally apply the same bidirectional objective to the overclustering branch on novel points, using its OT assignments over $\gamma C_{\mathrm{nov}}$ slots, and denote the resulting loss by $\mathcal L_{\mathrm{over}}$. The overall discovery loss is thus defined as $\mathcal L_{\mathrm{disc}}=\frac{1}{2}(\mathcal L_{\mathrm{ce}}+\mathcal L_{\mathrm{over}})$.

To further improve segmentation quality on the target domain, we add a Lovasz-Softmax regularization term \cite{berman2018lovasz} on the shared points. Let $\hat y_{v,n}^{t,h}=\arg\max_c \mathbf q_{v,n}^{t,h}(c)$ be the hard pseudo label converted from the soft target. The loss is defined as
\begin{equation}
\mathcal{L}_{\mathrm{lov}}
=
\frac{1}{H}\sum_{h=1}^{H}
\Big[
\ell_{\mathrm{lov}}(\mathbf{z}_{1,\Omega_t}^{t,h}, \hat{\mathbf{y}}_{2,\Omega_t}^{t,h})
+
\ell_{\mathrm{lov}}(\mathbf{z}_{2,\Omega_t}^{t,h}, \hat{\mathbf{y}}_{1,\Omega_t}^{t,h})
\Big],
\end{equation}
where $\ell_{\mathrm{lov}}(\cdot)$ denotes the Lovasz-Softmax loss~\cite{berman2018lovasz}. This term complements cross-entropy by directly optimizing region-level overlap and encourages sharper decision boundaries on the target domain.

When confidence-aware 2D-3D fusion is enabled, we further align the projected 2D features with the 3D point features through a cosine-similarity objective. For each view $v$, let $\mathcal V_t^v\subseteq\Omega_t$ denote the set of points that are both visible in the image branch and sufficiently reliable under the source model, and let $\omega_{v,n}^t$ be normalized confidence weights satisfying $\sum_{n\in\mathcal V_t^v}\omega_{v,n}^t=1$. The alignment loss is written as
\begin{equation}
\mathcal{L}_{\mathrm{2D\mbox{-}3D}}
=
\frac{1}{2}\sum_{v=1}^{2}
\sum_{n\in\mathcal{V}_t^v}
\omega_{v,n}^t
\left(
1-
\frac{
\langle \hat{\mathbf{u}}_{v,n}^t,\mathbf{f}_{v,n}^{t,\mathrm{3D}}\rangle
}{
\|\hat{\mathbf{u}}_{v,n}^t\|_2\,
\|\mathbf{f}_{v,n}^{t,\mathrm{3D}}\|_2
}
\right).
\end{equation}
This term encourages cross-modal semantic consistency while preventing unreliable points from dominating the fusion process.

The final optimization objective is
\begin{equation}
\mathcal{L}
=
\mathcal{L}_{\mathrm{disc}}
+
\lambda_{\mathrm{lov}}\mathcal{L}_{\mathrm{lov}}
+
\lambda_{\mathrm{align}}\mathcal{L}_{\mathrm{2D\mbox{-}3D}},
\end{equation}
where $\lambda_{\mathrm{lov}}$ and $\lambda_{\mathrm{align}}$ are the weights for the Lovasz regularization and the optional cross-modal alignment term, respectively. In practice, we set $\lambda_{\mathrm{lov}}=0.5$, and linearly warm up $\lambda_{\mathrm{align}}$ over the first 1000 steps as $\min(\mathrm{step}/1000, 1.0)$. When no image prior is used, $\lambda_{\mathrm{align}}$ is set to 0.
\begin{table*}[!t]
\centering
\tiny
\scriptsize
\renewcommand\arraystretch{1.0}
\setlength\tabcolsep{0.3pt}
\caption{The novel class discovery results on the SynLiDAR $\rightarrow$ SemanticKITTI setting with 19 classes. `Source' denotes the results obtained by source model only. `Full' denotes training under full supervision with ground-truth labels. The four groups represent the four splits in turn, and the gray values are the novel classes in each split. * indicates our re-implementation.
}
\label{tab:synlidar_to_semantickitti}
\begin{adjustbox}{width=\textwidth}
\renewcommand{\arraystretch}{0.3}
\begin{tabular}{c|ccccccccccccccccccc|ccc}
\toprule
\rotatebox{45}{Method} & \rotatebox{45}{bi.cle} & \rotatebox{45}{b.clst}  & \rotatebox{45}{build.}  & \rotatebox{45}{car}     & \rotatebox{45}{fence}   & \rotatebox{45}{mt.cle}  & \rotatebox{45}{m.clst}  & \rotatebox{45}{oth-g.} & \rotatebox{45}{oth-v.}  & \rotatebox{45}{park.}   & \rotatebox{45}{pers.}   & \rotatebox{45}{pole}    & \rotatebox{45}{road}    & \rotatebox{45}{sidew.}  & \rotatebox{45}{terra.}  & \rotatebox{45}{traff.}  & \rotatebox{45}{truck}   & \rotatebox{45}{trunk}   & \rotatebox{45}{veget.}  & \rotatebox{45}{Novel}   & \rotatebox{45}{Known} & \rotatebox{45}{All}  \\ \midrule

Full & 16.8 & 62.1 & 32.8 & 86.5 & 13.1 & 50.7 & 0.6 & 4.5 & 34.8 & 18.4 & 44.2 & 48.1 & 38.7 & 6.9 & 21.5 & 25.8 & 52.4 & 41.9 & 19.3 & - & - & 32.6 \\
Source   & 12.3    & 23.1    & -    & 77.6    & 32.9    & 15.5    & 41.6     & 9.1    & 7.4    & 44.2    & 18.6    & 39.5    & -    & -    & -    & 23.0    & 30.3    & 29.9    & -    & -        & 28.9 & - \\ \cmidrule(lr){2-23}
NOPS$^*$ & \textbf{5.3} & 18.4 & \cellcolor[HTML]{EFEFEF}{19.9} & 67.4 & 7.0 & 16.9 & 18.3 & \textbf{0.3} & 2.2 & \textbf{13.0} & \textbf{18.7} & 32.5 & \cellcolor[HTML]{EFEFEF}{{51.2}} & \cellcolor[HTML]{EFEFEF}{18.1} & \cellcolor[HTML]{EFEFEF}{{12.0}} & 8.0 & 20.0 & 21.1 & \cellcolor[HTML]{EFEFEF}{31.8} & \cellcolor[HTML]{EFEFEF}{26.2} & 17.8 & 20.1 \\
SNOPS$^* $ & 1.7 & 0.0 & \cellcolor[HTML]{EFEFEF}{40.9} & 11.2 & 7.2 & 4.6 & 9.4 & 0.0 & 0.0 & 12.8 & 9.5 & 31.6 & \cellcolor[HTML]{EFEFEF}{\textbf{55.3}} & \cellcolor[HTML]{EFEFEF}{1.2} & \cellcolor[HTML]{EFEFEF}{\textbf{33.7}} & 0.9 & 10.4 & 15.8 & \cellcolor[HTML]{EFEFEF}{27.5} & \cellcolor[HTML]{EFEFEF}{31.7} & 8.2 & 14.4 \\

Ours  & 3.0 & \textbf{37.1} & \cellcolor[HTML]{EFEFEF}{\textbf{50.3}} & \textbf{83.5} & \textbf{16.8} & \textbf{18.0} & \textbf{24.7} & 0.0 & \textbf{11.2} & 9.9 & 17.4 & \textbf{39.4} & \cellcolor[HTML]{EFEFEF}{ 42.8 } & \cellcolor[HTML]{EFEFEF}{\textbf{19.9}} & \cellcolor[HTML]{EFEFEF}{ 3.5 } & \textbf{26.9} & \textbf{31.4} & \textbf{32.8} & \cellcolor[HTML]{EFEFEF}{\textbf{51.4}} & \cellcolor[HTML]{EFEFEF}{ \textbf{33.6} } & \textbf{25.2} & \textbf{27.4} \\ \midrule

Full & 31.6 & 74.9 & 88.4 & 27.6 & 16.7 & 24.7 & 0.0 & 4.1 & 4.9 & 39.9 & 48.7 & 0.0 & 92.0 & 76.8 & 64.5 & 40.3 & 2.4 & 59.9 & 88.7 & - & - & 41.4 \\
Source   & 0.0    & 11.7    & 41.5    & -    & -    & 5.6    & 9.9     & 0.0    & -    & 2.7    & 9.6    & -    & 63.7    & 34.3    & 20.6   & 15.1    & -    & 14.4   & 60.6    & -        & 20.7 & - \\  \cmidrule(lr){2-23}
NOPS$^*$ & 0.0 & 46.6 & 43.3 & \cellcolor[HTML]{EFEFEF}{27.3} & \cellcolor[HTML]{EFEFEF}{1.3} & 9.1 & 1.7 & 0.0 & \cellcolor[HTML]{EFEFEF}{2.0} & 3.1 & 14.1 & \cellcolor[HTML]{EFEFEF}{18.5} & \textbf{58.9} & 32.9 & 21.3 & 0.6 & \cellcolor[HTML]{EFEFEF}{0.4} & 17.0 & 61.8 & \cellcolor[HTML]{EFEFEF}{9.9} & 22.2 & 18.9 \\
SNOPS$^* $ & 0.0 & 0.1 & \textbf{49.5} & \cellcolor[HTML]{EFEFEF}{26.2} & \cellcolor[HTML]{EFEFEF}{0.0} & 5.4 & 0.0 & 0.0 & \cellcolor[HTML]{EFEFEF}{2.8} & 0.0 & 8.0 & \cellcolor[HTML]{EFEFEF}{0.4} & 24.4 & 27.4 & 16.9 & 0.0 & \cellcolor[HTML]{EFEFEF}{1.3} & 2.0 & 61.7 & \cellcolor[HTML]{EFEFEF}{14.0} & 6.1 & 11.9 \\
Ours   & \textbf{9.3} & \textbf{52.9} & {47.8} & \cellcolor[HTML]{EFEFEF}{\textbf{30.9}} & \cellcolor[HTML]{EFEFEF}{\textbf{8.8}} & {\textbf{16.6}} & \textbf{10.3} & \textbf{1.7} & \cellcolor[HTML]{EFEFEF}{\textbf{4.7}} & \textbf{{8.3}} & \textbf{15.8} & \cellcolor[HTML]{EFEFEF}{\textbf{24.3}} & {58.7} & \textbf{36.1} & \textbf{27.6} & \textbf{12.8} & \cellcolor[HTML]{EFEFEF}{\textbf{3.3}} & {\textbf{19.8}} & \textbf{63.1} & \cellcolor[HTML]{EFEFEF}{\textbf{14.4}} & \textbf{27.2} & \textbf{23.8} \\ \midrule

Full & 33.0 & 11.9 & 90.5 & 93.6 & 39.1 & 47.0 & 1.1 & 0.0 & 46.9 & 37.3 & 2.2 & 56.9 & 92.1 & 76.0 & 64.3 & 31.1 & 67.6 & 19.3 & 89.6 & - & - & 47.3 \\
Source   & 3.4    & -    & 42.7    & 43.4    & 8.7    & 6.7    & -     & -     & 1.6    & 0.7    & -    & 27.2    & 55.4    & 32.1    & 17.4   & 11.1    & 11.6    & -   & 61.8    & -        & 23.1 & - \\  \cmidrule(lr){2-23}
NOPS$^* $ & \textbf{5.5} & \cellcolor[HTML]{EFEFEF}{1.6} & 44.5 & 34.8 & 8.2 & 11.4 & \cellcolor[HTML]{EFEFEF}{0.0} & \cellcolor[HTML]{EFEFEF}{0.0} & 4.4 & 9.4 & \cellcolor[HTML]{EFEFEF}{0.8} & 25.2 & 58.0 & 31.2 & 13.7 & 15.1 & 13.9 & \cellcolor[HTML]{EFEFEF}{2.8} & 56.9 & \cellcolor[HTML]{EFEFEF}{1.0} & 23.7 & 17.8 \\
SNOPS$^* $ & 0.0 & \cellcolor[HTML]{EFEFEF}{12.3} & 30.9 & 1.8 & 0.3 & 2.9 & \cellcolor[HTML]{EFEFEF}{0.0} & \cellcolor[HTML]{EFEFEF}{0.0} & 0.0 & 0.0 & \cellcolor[HTML]{EFEFEF}{0.2} & 19.1 & 50.7 & 31.4 & 10.2 & 0.0 & 2.6 & \cellcolor[HTML]{EFEFEF}{0.9} & 47.6 & \cellcolor[HTML]{EFEFEF}{2.7} & 14.1 & 11.1 \\
Ours & 4.0 & \cellcolor[HTML]{EFEFEF}{\textbf{13.8}} & \textbf{50.0} & \textbf{62.8} & \textbf{10.7} & \textbf{16.1} & \cellcolor[HTML]{EFEFEF}{\textbf{1.2}} & \cellcolor[HTML]{EFEFEF}{\textbf{0.7}} & \textbf{6.5} & \textbf{11.9} & \cellcolor[HTML]{EFEFEF}{\textbf{7.5}} & \textbf{37.2} & \textbf{63.4} & \textbf{38.9} & \textbf{30.6} & \textbf{18.9} & \textbf{25.4} & \cellcolor[HTML]{EFEFEF}{\textbf{11.5}} & \textbf{65.3} & \cellcolor[HTML]{EFEFEF}{\textbf{6.94}} & \textbf{31.6} & \textbf{25.1} \\ \midrule

Full & 0.0 & 81.0 & 90.2 & 93.6 & 37.0 & 0.0 & 0.0 & 0.7 & 50.7 & 0.0 & 49.9 & 58.0 & 90.6 & 72.8 & 63.1 & 1.9 & 71.0 & 60.2 & 89.4 & - & - & 47.9 \\
Source   & -  & 15.3    & 23.2    & 60.9    & 9.8    & -    & 18.6     & 0.3     & 5.2    & -    & 12.3    & 25.6    & 68.0    & 33.8    & 17.2   & -    & 14.6    & 12.2   & 57.4    & -    & 25.0 & - \\ \cmidrule(lr){2-23}
NOPS$^*$ & \cellcolor[HTML]{EFEFEF}{0.0} & 8.9 & 39.2 & 40.2 & 7.8 & \cellcolor[HTML]{EFEFEF}{0.0} & 2.9 & 0.0 & 0.2 & \cellcolor[HTML]{EFEFEF}{0.0} & 7.5 & 26.7 & 40.1 & 33.2 & 24.8 & \cellcolor[HTML]{EFEFEF}{0.0} & 13.0 & 13.8 & 64.0 & \cellcolor[HTML]{EFEFEF}{0.0} & 21.5 & 17.0 \\
SNOPS$^* $ & \cellcolor[HTML]{EFEFEF}{0.0} & 0.0 & 14.3 & 0.0 & 0.0 & \cellcolor[HTML]{EFEFEF}{0.0} & 0.0 & 0.0 & 0.0 & \cellcolor[HTML]{EFEFEF}{0.0} & 6.2 & 33.1 & 0.1 & 27.4 & 23.6 & \cellcolor[HTML]{EFEFEF}{0.7} & 3.8 & 8.8 & 64.1 & \cellcolor[HTML]{EFEFEF}{0.1} & 12.1 & 9.6 \\
Ours & \cellcolor[HTML]{EFEFEF}{0.0} & \textbf{28.6} & \textbf{38.1} & \textbf{66.3} & \textbf{10.5} & \cellcolor[HTML]{EFEFEF}{0.0} & \textbf{16.4} & \textbf{0.2} & \textbf{7.8} & \cellcolor[HTML]{EFEFEF}{\textbf{6.9}} & \textbf{19.7} & \textbf{35.6} & \textbf{59.3} & \textbf{39.3} & \textbf{31.2} & \cellcolor[HTML]{EFEFEF}{\textbf{9.0}} & \textbf{19.0} & \textbf{18.8} & \textbf{64.2} & \cellcolor[HTML]{EFEFEF}{\textbf{8.0}} & \textbf{30.33} & \textbf{24.8} \\ \midrule

\end{tabular}
\end{adjustbox}
 
\end{table*}

\section{Experiments}

\begin{table*}[htbp]
\caption{The novel class discovery results on the Synth4D $\rightarrow$ SemanticKITTI setting with 7 classes. 
}
\label{tab:synth4d_to_semantickitti}
\renewcommand{\arraystretch}{0.4}
\begin{tabular}{cc|ccccccc|ccc}
\midrule
                              &                          &                                        &                                       &                                        &                                        &                                        &                                        &                                        & \multicolumn{3}{c}{mIoU}                                                 \\
\multirow{-2}{*}{Split}       & \multirow{-2}{*}{Method} & \multirow{-2}{*}{vehicle}              & \multirow{-2}{*}{person}              & \multirow{-2}{*}{road}                 & \multirow{-2}{*}{sidewalk}             & \multirow{-2}{*}{terrain}              & \multirow{-2}{*}{manmade}              & \multirow{-2}{*}{vegetation}           & Novel                                  & Known          & All            \\ \midrule
                              & Full  & 77.03                                  & 0.00                                  & 20.29                                  & 19.35                                  & 21.07                                  & 77.61                                  & 81.28                                  & -                                      & -              & 42.37          \\
                              & Source & 75.50                                  & 14.35                                 & -                                      & -                                      & -                                      & 39.91                                  & 71.42                                  & -                                      & 50.29          & -              \\ \cmidrule(lr){2-12}
                              & NOPS$^*$   & 45.58                                  & 7.68                                  & \cellcolor[HTML]{EFEFEF}{44.70} & \cellcolor[HTML]{EFEFEF}10.76          & \cellcolor[HTML]{EFEFEF}14.75          & {38.84}                         & 64.92                                  & \cellcolor[HTML]{EFEFEF}{24.40} & 39.26          & 32.46          \\
                              & SNOPS$^*$  & 68.50                  &  \textbf{14.94}                        & \cellcolor[HTML]{EFEFEF} \textbf{58.85}              & \cellcolor[HTML]{EFEFEF} \textbf{30.12}          & \cellcolor[HTML]{EFEFEF} \textbf{34.36}           &   \textbf{52.16}                                &  \textbf{67.53}                  &  \cellcolor[HTML]{EFEFEF}\textbf{41.11}             &  \textbf{50.78}                             &   \textbf{46.64}             \\
\multirow{-5}{*}{\raisebox{10ex}{Synth4d$^0$}} & Ours   & \textbf{69.54}                         & {13.14}                        & \cellcolor[HTML]{EFEFEF}21.25          & \cellcolor[HTML]{EFEFEF}{18.15} & \cellcolor[HTML]{EFEFEF}{21.76} & 38.53                                  & {66.96}                         & \cellcolor[HTML]{EFEFEF}20.40          & {47.10} & {35.65} \\ \midrule
                              & Full  & 43.96                                  & 9.70                                  & 86.62                                  & 60.43                                  & 50.43                                  & 37.38                                  & 79.07                                  & -                                      & -              & 52.51          \\
                              & Source & -                                      & 19.67                                 & 71.62                                  & 25.51                                  & 19.72                                  & -                                      & 81.36                                  & -                                      & 43.58          & -              \\ \cmidrule(lr){2-12}
                              & NOPS$^*$   & \cellcolor[HTML]{EFEFEF}{27.96} & 0.14                                  & 70.26                                  & 23.18                                  & 20.80                                  & \cellcolor[HTML]{EFEFEF}33.73          & 69.08                                  & \cellcolor[HTML]{EFEFEF}{30.85} & 36.69          & 35.02          \\
                              & SNOPS$^*$  & \cellcolor[HTML]{EFEFEF} \textbf{37.76}              &  0.0                                     &  67.34                                  &   20.76                                     &   18.88                                & \cellcolor[HTML]{EFEFEF}  37.19             & \textbf{73.70}                                       & \cellcolor[HTML]{EFEFEF} \textbf{37.48}              &  37.14              &  36.52        \\
\multirow{-5}{*}{\raisebox{10ex}{Synth4d$^1$}} & Ours   & \cellcolor[HTML]{EFEFEF}10.11          & \textbf{10.23}                        & \textbf{72.24}                         & \textbf{25.14}                         & \textbf{24.87}                         & \cellcolor[HTML]{EFEFEF}\textbf{47.00} & {71.39}                         & \cellcolor[HTML]{EFEFEF}28.56          & \textbf{40.77} & \textbf{37.37} \\ \midrule
                              & Full  & 82.30                                  & 0.38                                  & 85.32                                  & 57.37                                  & 44.59                                  & 61.36                                  & 47.07                                  & -                                      & -              & 54.06          \\
                              & Source & 67.56                                  & -                                     & 62.98                                  & 24.18                                  & 27.43                                  & 81.28                                  & -                                      & -                                      & 52.69          & -              \\ \cmidrule(lr){2-12}
                              & NOPS$^*$   & 18.02                                  & \cellcolor[HTML]{EFEFEF} 0.38 & 67.50                                  & 20.27                                  & 27.22                                  & 52.23                                  & \cellcolor[HTML]{EFEFEF}31.98          & \cellcolor[HTML]{EFEFEF}16.18          & 37.05          & 31.07          \\
                              & SNOPS$^*$  & 42.94 & \cellcolor[HTML]{EFEFEF} \textbf{1.27}  &  69.02  & 21.84 & 28.87   & 54.40  & \cellcolor[HTML]{EFEFEF} 38.51  & \cellcolor[HTML]{EFEFEF} 19.89 & 43.41    &  36.69 \\
\multirow{-5}{*}{\raisebox{10ex}{Synth4d$^2$}} & Ours   & \textbf{74.83}                         & \cellcolor[HTML]{EFEFEF}0.33          & \textbf{72.17}                         & \textbf{22.60}                         & \textbf{30.85}                         & \textbf{56.73}                         & \cellcolor[HTML]{EFEFEF}\textbf{53.54} & \cellcolor[HTML]{EFEFEF}\textbf{26.94} & \textbf{51.44} & \textbf{44.44} \\ \midrule
\end{tabular}
\end{table*}

\begin{table*}[htbp]
\caption{The novel class discovery results on the Synth4D $\rightarrow$ nuScenes setting with 7 classes. 
}
\label{tab:synth4d_to_nusences}
\renewcommand{\arraystretch}{0.4}
\begin{tabular}{cc|ccccccc|ccc}
\midrule
                              &                          &                                        &                                       &                                        &                                        &                                        &                                        &                                        & \multicolumn{3}{c}{mIoU}                                                 \\
\multirow{-2}{*}{Split}       & \multirow{-2}{*}{Method} & \multirow{-2}{*}{vehicle}              & \multirow{-2}{*}{person}              & \multirow{-2}{*}{road}                 & \multirow{-2}{*}{sidewalk}             & \multirow{-2}{*}{terrain}              & \multirow{-2}{*}{manmade}              & \multirow{-2}{*}{vegetation}           & Novel                                  & Known          & All            \\ \midrule
                              & Full                     & 63.59                                  & 24.78                                 & 40.37                                  & 12.16                                  & 5.09                                   & 74.83                                  & 61.24                                  & -                                      & -              & 40.29          \\
                              & Source                   & 66.80                                  & 24.21                                 & -                                      & -                                      & -                                      & 76.91                                  & 60.89                                  & -                                      & 57.20          & -              \\ \cmidrule(lr){2-12} 
                              & NOPS$^*$                     & 58.37                                  & 18.15                        & \cellcolor[HTML]{EFEFEF}43.35          & \cellcolor[HTML]{EFEFEF}6.72           & \cellcolor[HTML]{EFEFEF}\textbf{8.41}           & 68.29                                  & 49.87                                  & \cellcolor[HTML]{EFEFEF}19.49          & 48.67          & 36.17          \\
                              & SNOPS$^*$                    & 51.11                         & 2.15                                & \cellcolor[HTML]{EFEFEF}\textbf{45.65} & \cellcolor[HTML]{EFEFEF}\textbf{12.75} & \cellcolor[HTML]{EFEFEF}3.12 & 58.48                                  & 56.07                         & \cellcolor[HTML]{EFEFEF}\textbf{20.51} & 41.95          & 32.76 \\
\multirow{-5}{*}{\raisebox{10ex}{Synth4d$^0$}} & Ours                     & \textbf{64.72}                                  & \textbf{20.48}                                 & \cellcolor[HTML]{EFEFEF}37.15          & \cellcolor[HTML]{EFEFEF}8.77           & \cellcolor[HTML]{EFEFEF}3.90           & \textbf{72.94}                         & \textbf{62.91}                                  & \cellcolor[HTML]{EFEFEF}16.61          & \textbf{55.26} & \textbf{38.70}          \\ \midrule
                              & Full                     & 23.42                                  & 16.26                                 & 91.37                                  & 45.99                                  & 31.78                                  & 34.49                                  & 55.75                                  & -                                      & -              & 46.38          \\
                              & Source                   & -                                      & 25.26                                 & 75.03                                  & 22.85                                  & 11.93                                  & -                                      & 73.57                                  & -                                      & 41.72          & -              \\ \cmidrule(lr){2-12} 
                              & NOPS$^*$                     & \cellcolor[HTML]{EFEFEF}22.16          & 6.61                                  & 78.00                                  & 22.83                                  & 12.46                                  & \cellcolor[HTML]{EFEFEF}42.06          & 52.08                                  & \cellcolor[HTML]{EFEFEF}32.11          & 34.40          & 33.74          \\
                              & SNOPS$^*$                    & \cellcolor[HTML]{EFEFEF}18.22          & 5.79                                   & 74.14                                  & 23.74                                  & 9.12                                  & \cellcolor[HTML]{EFEFEF}37.11          & {48.35}                         & \cellcolor[HTML]{EFEFEF}{27.67} & 32.23          & 30.92          \\
\multirow{-5}{*}{\raisebox{10ex}{Synth4d$^1$}} & Ours                     & \cellcolor[HTML]{EFEFEF}\textbf{25.91} & \textbf{17.09}                        & \textbf{89.87}                         & \textbf{44.84}                         & \textbf{31.18}                         & \cellcolor[HTML]{EFEFEF}\textbf{46.92} & \textbf{54.07}                                  & \cellcolor[HTML]{EFEFEF}\textbf{36.42}          & \textbf{47.41} & \textbf{44.27} \\ \midrule
                              & Full                     & 67.67                                  & 0.54                                  & 91.57                                  & 48.00                                  & 26.54                                  & 77.75                                  & 46.53                                  & \cellcolor[HTML]{EFEFEF}-              & -              & 51.23          \\
                              & Source                   & 63.45                                  & -                                     & 65.94                                  & 18.49                                  & 12.49                                  & 78.05                                  & -                                      & -                                      & 47.68          & -              \\ \cmidrule(lr){2-12} 
                              & NOPS$^*$                     & 50.80                                  & \cellcolor[HTML]{EFEFEF}0.43          & 71.12                                  & 18.30                                  & 12.52                                  & 65.05                                  & \cellcolor[HTML]{EFEFEF}29.60          & \cellcolor[HTML]{EFEFEF}15.02          & 43.56          & 35.40          \\
                              & SNOPS$^*$                    & 42.94                                  & \cellcolor[HTML]{EFEFEF}\textbf{1.27} & 69.02                                  & 21.84                                  & \textbf{28.87}                         & 54.40                                  & \cellcolor[HTML]{EFEFEF}38.51          & \cellcolor[HTML]{EFEFEF}19.89          & 43.41          & 36.69          \\
\multirow{-5}{*}{\raisebox{10ex}{Synth4d$^2$}} & Ours                     & \textbf{64.59}                         & \cellcolor[HTML]{EFEFEF}1.04          & \textbf{90.90}                         & \textbf{46.08}                         & 24.94                                  & \textbf{74.16}                         & \cellcolor[HTML]{EFEFEF}\textbf{40.56} & \cellcolor[HTML]{EFEFEF}\textbf{20.80} & \textbf{60.13} & \textbf{48.90} \\ \midrule
\end{tabular}
\end{table*}

\subsection{Experimental Setup}
\noindent \textbf{Source and target datasets.} 
We use SynLiDAR~\cite{SynLiDAR} and Synth4D~\cite{GIPSO} as the source datasets for pre-training, and adapt the pre-trained models to the real-world target datasets SemanticKITTI~\cite{semantickitti} and nuScenes~\cite{nuscenes}. For both target datasets, we follow the official validation split. For SemanticKITTI, this corresponds to Sequence 08, which contains 4,071 LiDAR scans at 10 Hz. For nuScenes, we use the official validation split with 150 sequences and 6,000 LiDAR scans, where 40 scans are sampled from each scene at 2 Hz.

\noindent \textbf{Baselines.} 
We select several novel category discovery methods, including NOPS~\cite{NOPS} and SNOPS~\cite{SNOPS}, and adapt them as baselines for the GCD-TTA setting. To make these methods compatible with our setting, we apply the required modifications, in particular using the pretrained source model to generate pseudo labels for adaptation. For SNOPS, the auxiliary network alignment stage does not converge with the original SGD optimizer in our experiments, whereas AdamW~\cite{adamw} yields stable results, and is therefore used in our implementation. Other methods, including DASL~\cite{DASL}, Li et al.~\cite{CaRR}, and Kim et al.~\cite{Kim_2026_WACV}, are excluded due to their code is unavailable or incomplete in core parts.

\noindent \textbf{Implementation details.} 
We use MinkowskiNet~\cite{Minkowski} as the backbone for LiDAR segmentation. In total, we pre-train 10 source models across three settings, namely four splits for SynLiDAR $\rightarrow$ SemanticKITTI, three splits for Synth4D $\rightarrow$ SemanticKITTI, and three splits for Synth4D $\rightarrow$ nuScenes. The models are trained for 200 epochs on SynLiDAR $\rightarrow$ SemanticKITTI and 100 epochs on the other two settings. We use SGD with momentum 0.9, weight decay 0.0001, and batch size 16. All source pre-training experiments are conducted on 8 NVIDIA RTX 4090 GPUs.
During adaptation, we perform online test-time adaptation by traversing the unlabeled target validation set only once with a batch size of 1. The optimizer and other hyper-parameters are kept the same as in pre-training. For the GCD procedure, we follow~\cite{GIPSO,swav} and adopt the Sinkhorn-Knopp~\cite{cuturi2013sinkhorn} algorithm with 3 Sinkhorn iterations. All adaptation experiments are performed on a single NVIDIA RTX 4090 GPU.

\noindent \textbf{Evaluation metrics.}
For evaluation, we adopt the strict Hungarian matching algorithm~\cite{Hungarian} following GCD~\cite{gcd}. Hungarian matching is applied only to category discovery, since their predicted indices are ambiguous under clustering. We report the final performance of the mean Intersection over Union (mIoU) for both known and novel categories.

\subsection{Quantitative Comparison}
\noindent \textbf{Synthetic-to-real setting.} 
Tabs.~\ref{tab:synlidar_to_semantickitti},~\ref{tab:synth4d_to_semantickitti}, and~\ref{tab:synth4d_to_nusences} report the mIoU results on Novel, Known, and All classes. Overall, our method performs strongly and consistently across all three cross-domain settings. On SynLiDAR $\rightarrow$ SemanticKITTI, it achieves the best Novel, Known, and All mIoU on all four splits. On Synth4D $\rightarrow$ SemanticKITTI, it remains competitive and obtains the best All mIoU on two of the three splits. On Synth4D $\rightarrow$ nuScenes, it achieves the best All mIoU on all three splits, together with the best Novel mIoU on two splits. Overall, our method ranks first in All mIoU on 9 of the 10 splits, with up to 12.2 mIoU gain over the strongest re-implemented baseline, confirming the benefit of combining 3D geometric priors with 2D-3D multimodal semantic priors.

\begin{table}[htbp]
\caption{Real-to-real setting (Split 0).}
\label{tab:real_to_real}
\resizebox{\linewidth}{!}{
\begin{tabular}{lccc|ccc}
\hline
\multirow{2}{*}{Method} & \multicolumn{3}{c|}{SemanticKITTI$\rightarrow$nuScenes} & \multicolumn{3}{c}{nuScenes$\rightarrow$SemanticKITTI} \\ \cline{2-7} 
                        & Novel           & Known          & All          & Novel          & Known          & All          \\ \hline
NOPS                    & 30.15           & 28.47          & 29.19        & 26.67          & 29.08          & 28.04        \\
SNOPS                   & 24.04           & 19.79          & 21.61        & 39.21          & 41.42          & 40.47        \\
Ours                    & \textbf{46.72}  & \textbf{38.41} & \textbf{41.97}& \textbf{42.07}& \textbf{42.38} & \textbf{42.25}\\ \hline
\end{tabular}
}
\end{table}

\noindent \textbf{Real-to-real setting.}
Tab.~\ref{tab:real_to_real} reports the mIoU results on Novel, Known, and All classes for split 0 in the real-to-real setting. Overall, our method performs best and consistently across both transfer directions. On SemanticKITTI $\rightarrow$ nuScenes, it achieves the highest Novel, Known, and All mIoU, surpassing the stronger baseline by 16.57, 9.94, and 12.78 points, respectively. On nuScenes $\rightarrow$ SemanticKITTI, it again obtains the best results on all three metrics, with gains of 2.86 in Novel mIoU, 0.96 in Known mIoU, and 1.78 in All mIoU over the strongest baseline. These results show that our method generalizes effectively not only in synthetic-to-real transfer but also in real-to-real adaptation, demonstrating robust cross-domain transferability across different LiDAR datasets.

\subsection{Analysis of von Neumann Entropy} 
Following~\cite{tang2026bures}, we analyze the von Neumann entropy of feature representations. Given $X \in \mathbb{R}^{N \times d}$, we apply row-wise $\ell_2$ normalization to obtain $\tilde{X}$ and compute the autocorrelation matrix~\cite{scholkopf2002learning} as $\mathcal{A}=\frac{1}{N}\tilde{X}^{\top}\tilde{X}$. Let ${\lambda_i}_{i=1}^{d}$ be the eigenvalues of $\mathcal{A}$, normalized as $p_i=\lambda_i/\sum_j\lambda_j$. The entropy is defined as $\hat{H}(\mathcal{A})=-\sum_i p_i\log p_i$, and the effective rank is computed as $e^{\hat{H}(\mathcal{A})}$, where larger values indicate a more expressive and less collapsed feature space.

\begin{figure}[htbp]
    \centering
    \includegraphics[width=0.75\linewidth]{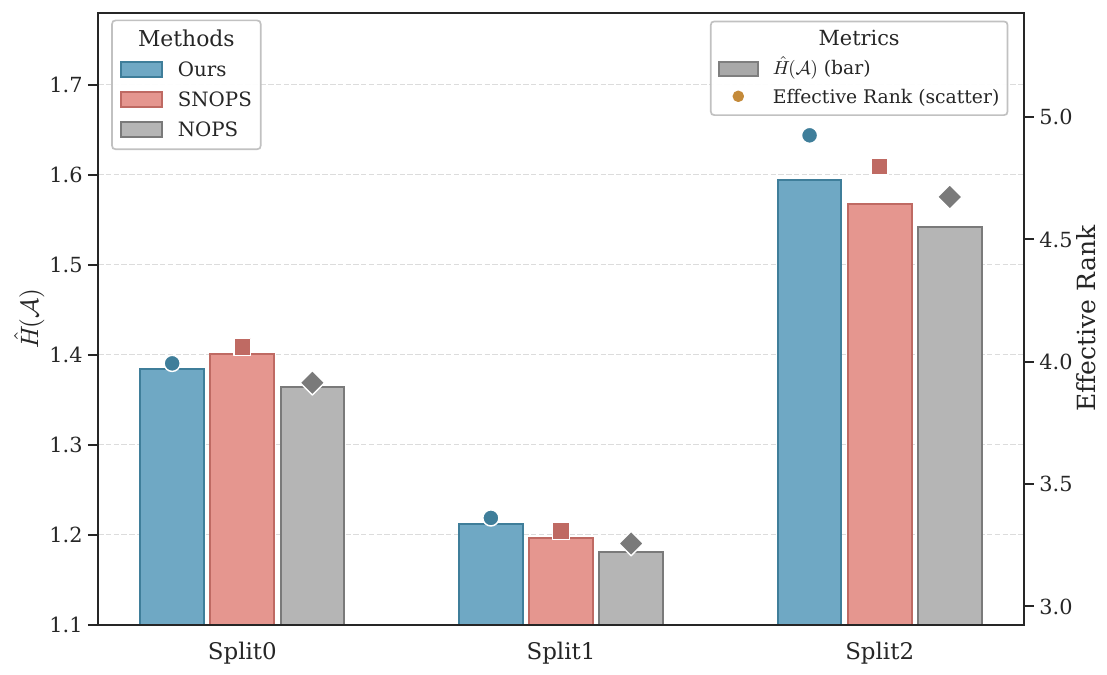}
    \caption{Comparison between $\hat H(\mathcal{A})$ and effective rank $e^{\hat H(\mathcal{A})}$. The count of the largest eigenvalues necessary to account for 99\% of the total eigenvalue energy serves as a surrogate for the rank.}
    \label{fig:H_effective_rank}
\end{figure}

As shown in Fig.~\ref{fig:H_effective_rank}, our method consistently achieves higher von Neumann entropy and effective rank than NOPS across all splits. It also outperforms SNOPS on Split 1 and Split 2 while remaining comparable on Split 0. These results indicate that our method preserves a less collapsed feature space with improved representation capacity.

\begin{figure}[t]
  \centering
\subfloat[SNOPS]
{\includegraphics[width=0.225\textwidth]{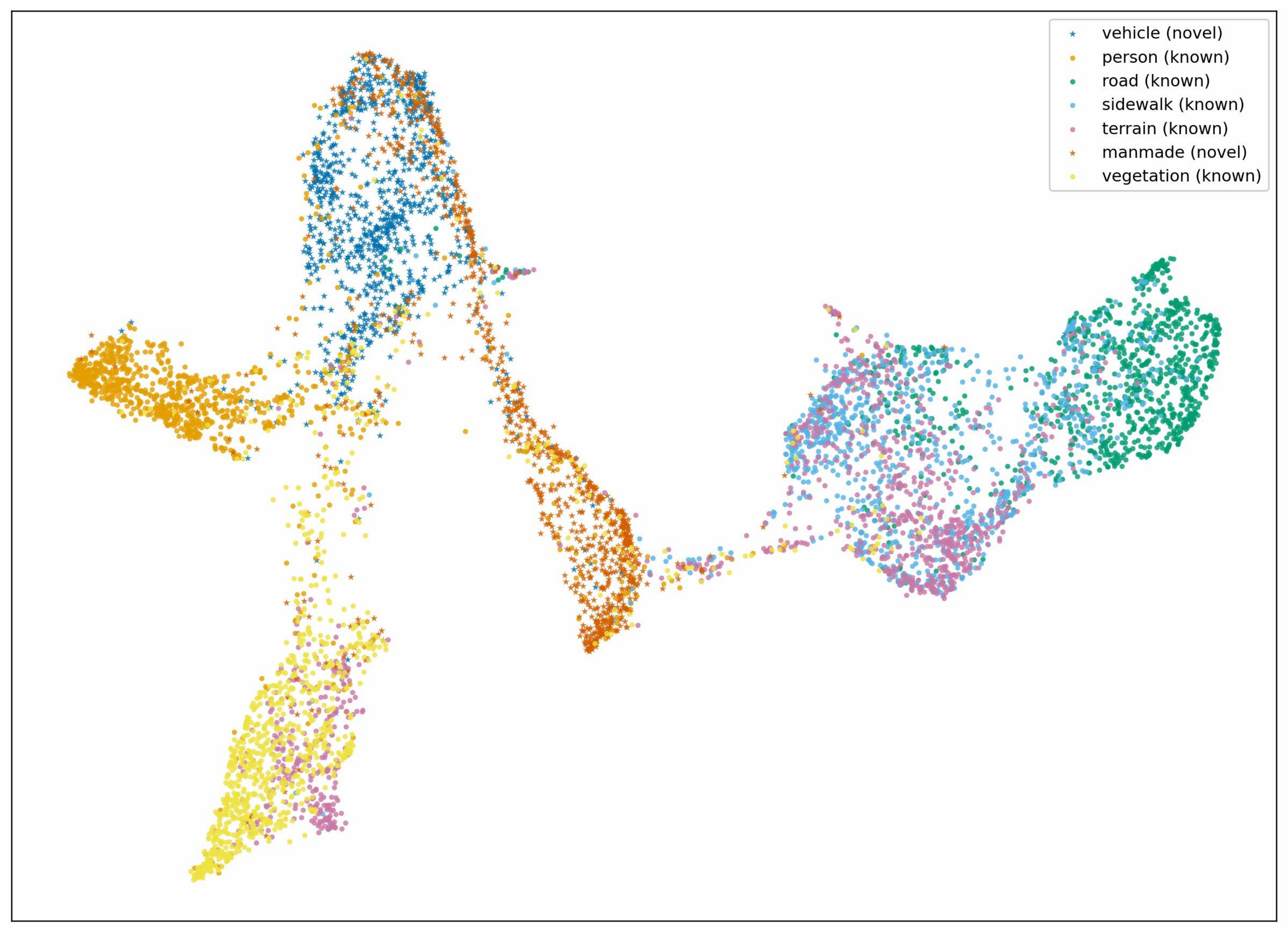}\label{fig:umap_snops}}
\quad
\subfloat[Ours]
{\includegraphics[width=0.225\textwidth]{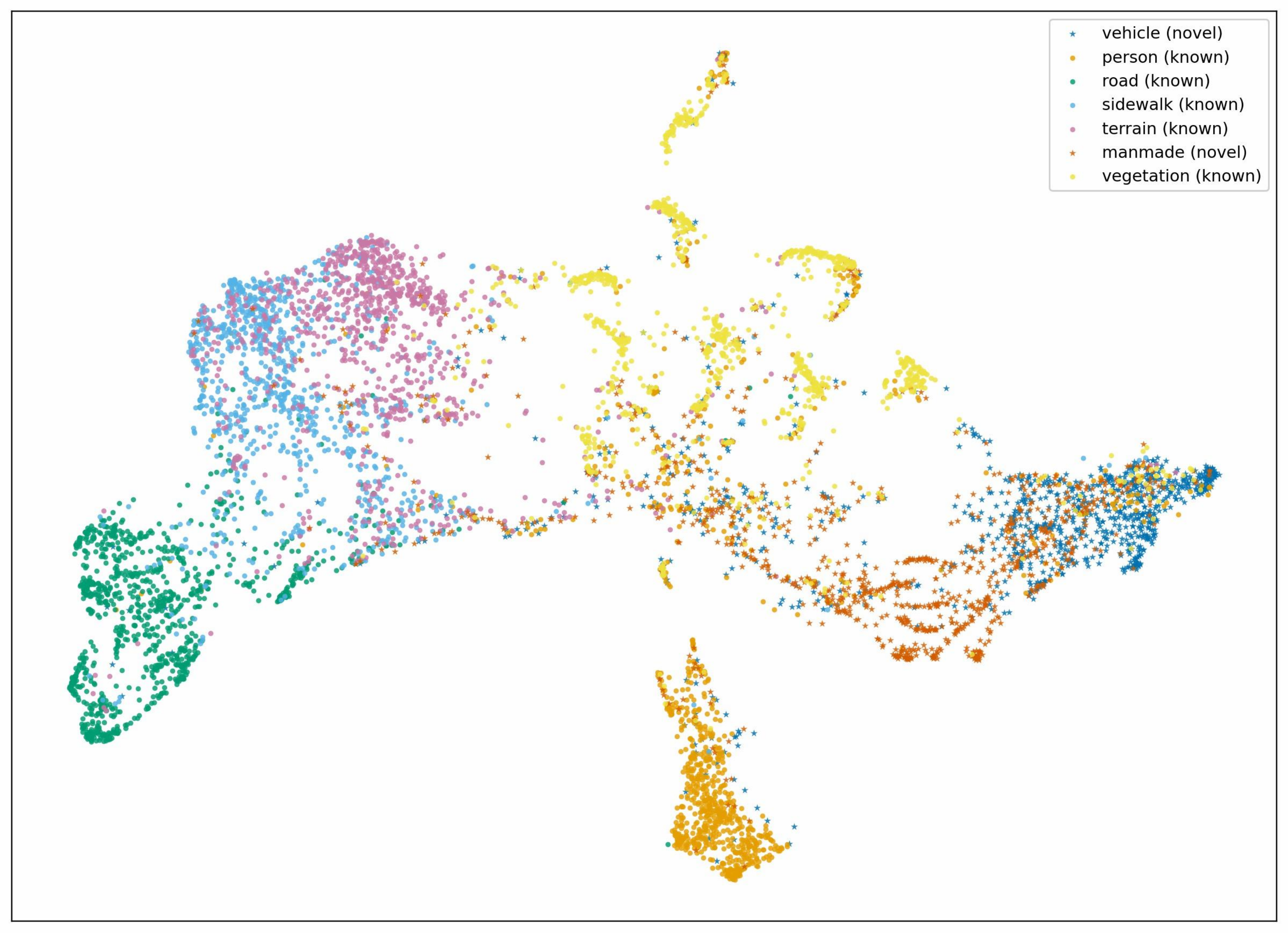}\label{fig:umap_ours}}
\quad
\caption{Comparison of UMAP visualizations between SNOPS and our method on the nuScenes dataset (Split 1).}
  \label{fig:umap}
\end{figure}

\subsection{Qualitative Analysis}
\noindent \textbf{UMAP visualization analysis.} We visualize feature distributions of our method and SNOPS on Split 1 of Synth4D $\rightarrow$ nuScenes using UMAP~\cite{UMAP}. Since online GCD-TTA remains highly challenging, neither method achieves fully separated clusters. Nevertheless, Fig.~\ref{fig:umap} reveals clear differences between them. Our method provides better separation for known classes, including \textcolor[HTML]{35b190}{\textit{road}}, \textcolor[HTML]{74c1ed}{\textit{sidewalk}}, and \textcolor[HTML]{d591b7}{\textit{terrain}}, whereas SNOPS exhibits stronger class mixing, with \textit{sidewalk} overlapping with both \textcolor[HTML]{35b190}{\textit{road}} and \textcolor[HTML]{d591b7}{\textit{terrain}}. Moreover, \textcolor[HTML]{d591b7}{\textit{terrain}} and \textcolor[HTML]{f2e864}{\textit{vegetation}} are more entangled in SNOPS. Although our \textcolor[HTML]{dc7b2e}{\textit{manmade}} and \textcolor[HTML]{f2e864}{\textit{vegetation}} features are less compact, the reduced inter-class overlap indicates more discriminative representations.

\begin{figure}[!t]
    \centering
    \includegraphics[width=\linewidth]{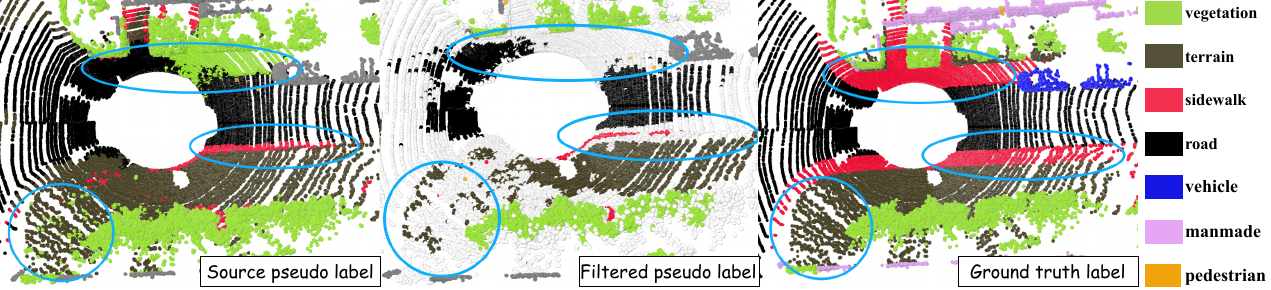}
    \caption{Pseudo label visualization (Split 0).}
    \label{fig:vis}
\end{figure}

\noindent \textbf{Qualitative comparison of pseudo labels.}
As shown in Fig.~\ref{fig:vis}, we visualize the source-model pseudo labels, filtered pseudo labels, and ground-truth labels for qualitative comparison. The source pseudo labels contain substantial noise, particularly among visually similar categories such as \textit{sidewalk}, \textit{terrain}, and \textit{road}, as highlighted by the blue circles. For instance, the upper elliptical region contains many incorrectly assigned points in the source predictions. After applying our filtering strategy, these unreliable predictions are effectively removed, producing cleaner pseudo labels that better align with the ground truth. Similar improvements are observed in other regions, where noisy predictions are suppressed while structural semantics are preserved. These results demonstrate that our method effectively filters erroneous pseudo labels and mitigates the impact of noisy supervision during optimization.

\begin{table}[!t]
\caption{Ablation on different numbers of 2D image views on Synth4D $\rightarrow$ nuScenes setting. F, FL, FR, B, BL, and BR denote FRONT, FRONT\_LEFT, FRONT\_RIGHT, BACK, BACK\_LEFT, and BACK\_RIGHT, respectively.}
\label{tab:dino_direction}
\resizebox{\linewidth}{!}{
\begin{tabular}{ccccccc}
\hline
\multirow{2}{*}{F} & \multirow{2}{*}{\makecell{FL \\ FR}} & \multirow{2}{*}{B} & \multirow{2}{*}{\makecell{BL \\ BR}} & \multicolumn{3}{c}{mIoU\textsubscript{all}} \\ \cline{5-7}
                   &                                        &                    &                                          & Split 0 & Split 1 & Split 2 \\ \hline
                   &                                        &                    &                                          & 34.78  & 38.83  & 41.22  \\ \cline{1-7}
$\checkmark$       &                                        &                    &                                          & 32.03 (\textcolor{BrickRed}{-2.75})  & 38.01 (\textcolor{BrickRed}{-0.82}) & 42.99 (\textcolor{ForestGreen}{\textbf{+1.77}}) \\
$\checkmark$       & $\checkmark$                           &                    &                                          & 33.75 (\textcolor{BrickRed}{-1.03})  & 39.88 (\textcolor{ForestGreen}{\textbf{+1.05}}) & 46.71 (\textcolor{ForestGreen}{\textbf{+5.49}}) \\
$\checkmark$       & $\checkmark$                           & $\checkmark$       & $\checkmark$                             & \textbf{38.70} (\textcolor{ForestGreen}{\textbf{+3.92}}) & \textbf{44.27} (\textcolor{ForestGreen}{\textbf{+5.44}}) & \textbf{48.90} (\textcolor{ForestGreen}{\textbf{+7.68}}) \\ \cline{1-7}
\multirow{2}{*}{F} & \multirow{2}{*}{\makecell{FL \\ FR}} & \multirow{2}{*}{B} & \multirow{2}{*}{\makecell{BL \\ BR}}     & \multicolumn{3}{c}{mIoU\textsubscript{novel}} \\ \cline{5-7}
                   &                                        &                    &                                          & Split 0 & Split 1 & Split 2 \\ \cline{1-7}
                   &                                        &                    &                                          & 16.71  & 31.51  & 13.32  \\ \cline{1-7}
$\checkmark$       &                                        &                    &                                          & 14.89 (\textcolor{BrickRed}{-1.82})  & 38.13 (\textcolor{ForestGreen}{\textbf{+6.62}}) & 19.10 (\textcolor{ForestGreen}{\textbf{+5.78}}) \\
$\checkmark$       & $\checkmark$                           &                    &                                          & \textbf{17.41} (\textcolor{ForestGreen}{\textbf{+0.70}})  & \textbf{38.48} (\textcolor{ForestGreen}{\textbf{+6.97}}) & \textbf{25.63} (\textcolor{ForestGreen}{\textbf{+12.31}}) \\
$\checkmark$       & $\checkmark$                           & $\checkmark$       & $\checkmark$                             & 16.61 (\textcolor{BrickRed}{-0.10}) & 36.42 (\textcolor{ForestGreen}{\textbf{+4.91}}) & 20.80 (\textcolor{ForestGreen}{\textbf{+7.48}}) \\ \hline
\end{tabular}
}
\end{table}

\subsection{Ablation Study}
\noindent \textbf{Different numbers of 2D image views.} 
Tab.~\ref{tab:dino_direction} shows that mIoU\textsubscript{all} and mIoU\textsubscript{novel} exhibit different responses to the number of 2D views. For mIoU\textsubscript{all}, the front view alone produces unstable results and even degrades performance on Split 0 and Split 1, indicating insufficient semantic coverage and biased feature learning. As more views are introduced, complementary image priors improve the representation, with all six views achieving the best mIoU\textsubscript{all} across all splits. For mIoU\textsubscript{novel}, additional 2D priors provide larger improvements, suggesting that semantic cues from foundation models help distinguish ambiguous target points beyond geometric evidence. However, the three front-facing views achieve the highest mIoU\textsubscript{novel}, while adding rear views slightly reduces performance on some splits. This may be because novel class discovery relies more on semantic consistency than coverage: extra views can introduce unreliable cues caused by occlusion, sparsity, and domain gaps between Synth4D and nuScenes. Overall, multi-view 2D priors are beneficial, but high-quality semantic guidance is more important than simply increasing the number of views.

\begin{table}[t]
\caption{Ablation study on whether to adopt adaptive kNN in the SynLiDAR $\rightarrow$ SemanticKITTI setting.}
\label{tab:w/o_adapt_knn}
\resizebox{\linewidth}{!}{
    \begin{tabular}{lcccc}
    \hline
                  & \multicolumn{4}{c}{mIoU}                                  \\ \cline{2-5}
                  & Split 0       & Split 1       & Split 2       & Split 3       \\ \hline
    w/o adapt kNN & 21.8         & 24.1         & 24.8         & 23.6         \\
    w/ adapt kNN  & 27.4 (\textcolor{ForestGreen}{\textbf{+5.6}})  & 23.8 (\textcolor{BrickRed}{-0.3})  & 25.1 (\textcolor{ForestGreen}{\textbf{+0.3}})  & 24.8 (\textcolor{ForestGreen}{\textbf{+1.2}}) \\ \hline
    \end{tabular}
}
\end{table}

\noindent \textbf{Effect of adaptive kNN.}
Tab.~\ref{tab:w/o_adapt_knn} shows the effect of adaptive kNN on SynLiDAR $\rightarrow$ SemanticKITTI. Using adaptive kNN improves the performance on three of the four splits, with gains of $+5.6$, $+0.3$, and $+1.2$ mIoU on Split 0, Split 2, and Split 3, respectively. Although it causes a slight drop of $0.3$ mIoU on Split 1, the overall trend suggests that adaptive kNN generally provides more suitable neighborhood relations under domain shift and is beneficial to segmentation performance.

\begin{figure}[!t]
    \centering
    \includegraphics[width=1.0\linewidth]{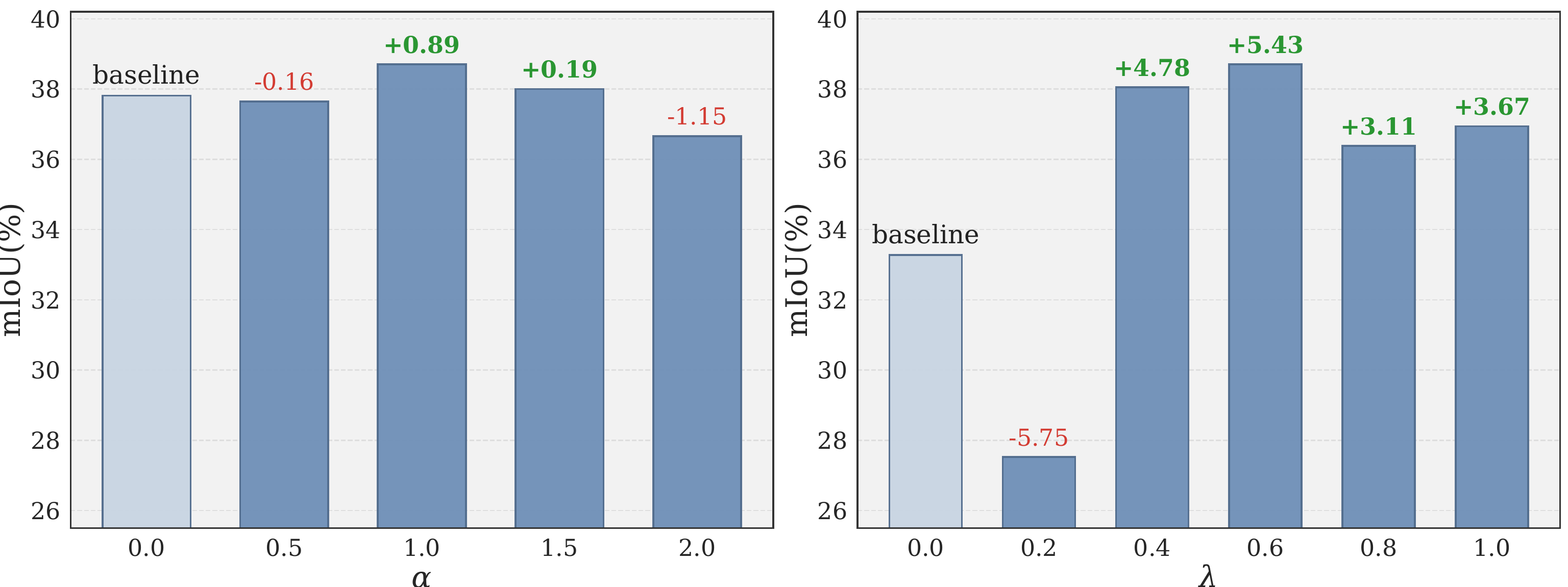}
    \caption{Impact of different hyper-parameters on the Synth4D $\rightarrow$ nuScenes setting (Split 0).}
    \label{fig:hyper-parameters_ablation}
\end{figure}
\noindent \textbf{Hyper-parameter analysis.}
Fig.~\ref{fig:hyper-parameters_ablation} shows the effects of two hyper-parameters: the adaptive-kNN exponent $\alpha$ (left) and the DINO fusion strength $\lambda$ (right). For adaptive kNN, $\alpha=1.0$ achieves the best mIoU of 38.70, outperforming the fixed-k baseline ($\alpha=0$) with 37.81. This suggests that moderate range-aware neighborhood adaptation effectively captures LiDAR density variation, while insufficient or excessive adaptation may degrade neighborhood stability, particularly for sparse distant points.

For DINO fusion, introducing 2D semantic cues consistently improves performance. The mIoU increases from 33.27 without fusion to 38.70 at $\lambda=0.6$, showing that 2D features compensate for limited 3D semantics. The exception is $\lambda=0.2$, which reduces performance to 27.52, likely because weak fusion fails to provide reliable semantic correction while disturbing the original 3D representation. With moderate fusion strengths ($0.4 \sim 1.0$), performance improves consistently, validating the effectiveness of confidence-aware cross-modal integration.

\section{Conclusion}
In this paper, we investigate generalized category discovery under test-time adaptation for 3D point cloud segmentation. To improve both domain adaptation and novel category discovery, we combine 3D geometric priors with 2D-3D multimodal semantic priors. Experiments on three cross-domain benchmarks show that the proposed method achieves consistent improvements, obtaining the best overall mIoU on 9 of the 10 splits. The ablation results further highlight the value of multi-view semantic priors, while the additional analyses suggest that our method learns features with better separation and less collapse. Although the current framework still assumes that the number of novel categories is known in advance and cannot fully avoid partial degradation on known classes compared with source-only training, the results indicate that integrating geometric and semantic priors is an effective direction for GCD-TTA in 3D LiDAR segmentation.

\noindent \textbf{Limitations.} 
Our method still has some limitations. In particular, it assumes prior knowledge of the number of novel categories, which may not be available in real-world applications. Moreover, although it alleviates the performance drop on known categories, partial degradation still exists. An important direction for future work is to develop category-number-free discovery strategies, further improve the preservation of known-category performance, and extend the proposed framework toward continual and incremental learning scenarios.

\begin{acks}
This work was supported in part by National Key R\&D Program of China (2024YFB3908500, 2024YFB3908502), NSFC Excellent Young Scientists Fund Program (Overseas), Guangdong Provincial Outstanding Young Talents Program, Guangdong Basic and Applied Basic Research Foundation (2023B1515120026), Shenzhen Science and Technology Program (KJZD20240903100022028), and Scientific Development Funds from Guangdong Provincial Key Laboratory of Visual Media and Multidimensional Intelligence.
\end{acks}

\bibliographystyle{ACM-Reference-Format}
\bibliography{a-main-bib}

\end{document}